\documentclass{egpubl}
\usepackage{eurovis2024}
\usepackage{enumitem}
\usepackage{color, soul}
\usepackage{algorithm2e}
\usepackage{subcaption}
\usepackage{setspace}

\SpecialIssueSubmission    

\CGFccby

\usepackage[T1]{fontenc}
\usepackage{dfadobe}  

\usepackage{cite}  
\BibtexOrBiblatex
\electronicVersion
\PrintedOrElectronic
\ifpdf \usepackage[pdftex]{graphicx} \pdfcompresslevel=9
\else \usepackage[dvips]{graphicx} \fi

\usepackage{egweblnk}

\usepackage{epstopdf}
\usepackage{dirtytalk}
\usepackage{booktabs}
\usepackage{makecell}

\title[Visual Analytics for Fine-grained Text Classification Models and Datasets]%
      {\vspace{-0.2in} Visual Analytics for Fine-grained Text Classification Models and Datasets}

\author[M. Battogtokh et al.]
{\parbox{\textwidth}{
    \centering 
    \vspace{-0.2in}M. Battogtokh\thanks{\vspace{-0.2in}Corresponding author: munkhtulga.battogtokh@kcl.ac.uk}$^{1}$\orcid{0000-0001-7562-414X},
    Y. Xing$^{1}$\orcid{0000-0003-1521-6616},
    C. Davidescu$^{2}$\orcid{0009-0003-8493-4563},
    A. Abdul-Rahman$^{1}$\orcid{0000-0002-6257-876X},
    M. Luck$^{1}$\orcid{0000-0002-0926-2061},
    and R. Borgo$^{1}$\orcid{0000-0003-2875-6793} 
}
        \\
{\parbox{\textwidth}{
    \centering \vspace{-0.2in}
    $^1$King's College London, United Kingdom\\
    $^2$ContactEngine, United Kingdom
       }
}
\vspace{-5ex}
}

\SetKwComment{Comment}{/* }{ */}

\begin{document}
\teaser{
 \includegraphics[width=0.9\linewidth]{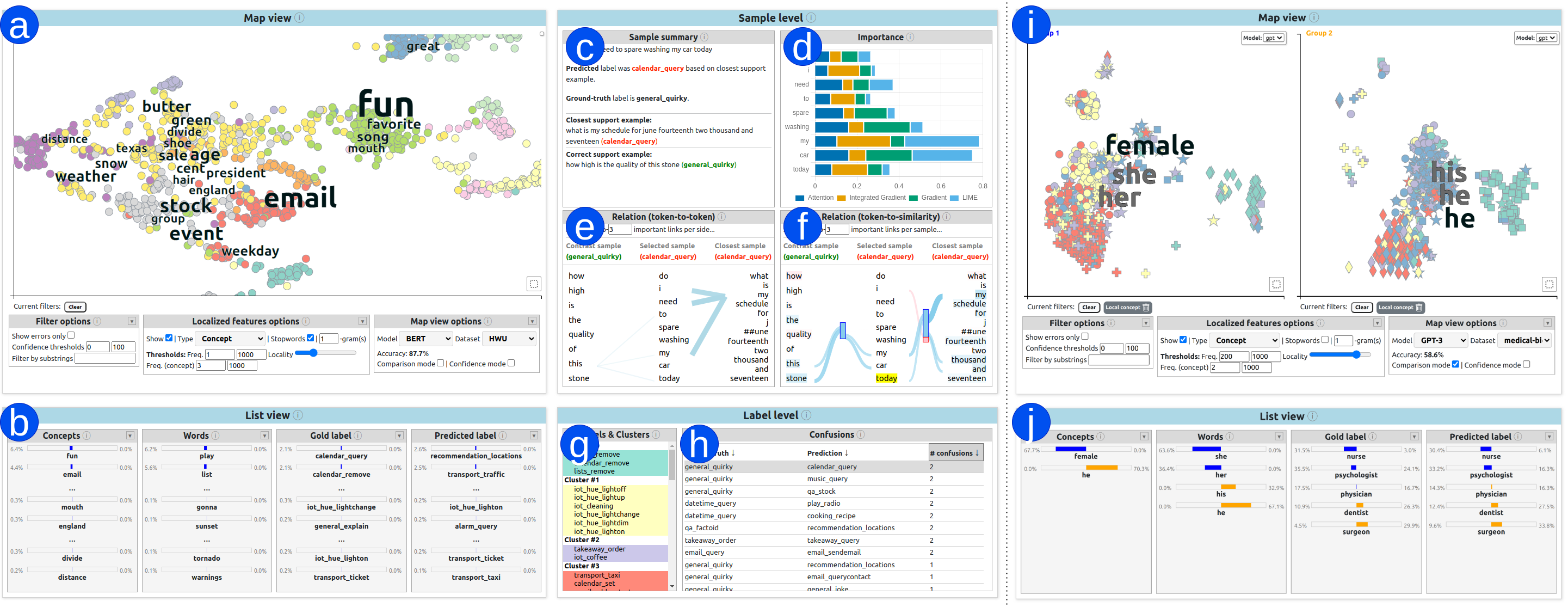}
 \centering
  \caption{User interface of SemLa (\textbf{Sem}antic \textbf{La}ndscape): Map view (a), List view (b), Sample-level consisting of natural language summary (c), Visually Integrated Feature Importance view (d), token-to-token and token-to-similarity relation graphs (e, f), Label-level comprising label-cluster list (g), confusion table (h), and the Map and List views in comparison mode (i and j).}
\label{fig:teaser}
}

\maketitle
\begin{abstract}
In natural language processing (NLP), text classification tasks are increasingly fine-grained, as datasets are fragmented into a larger number of classes that are more difficult to differentiate from one another. As a consequence, the semantic structures of datasets have become more complex, and model decisions more difficult to explain. Existing tools, suited for coarse-grained classification, falter under these additional challenges. In response to this gap, we worked closely with NLP domain experts in an iterative design-and-evaluation process to characterize and tackle the growing requirements in their workflow of developing fine-grained text classification models. The result of this collaboration is the development of SemLa, a novel Visual Analytics system tailored for 1) dissecting complex semantic structures in a dataset when it is spatialized in model embedding space, and 2) visualizing fine-grained nuances in the meaning of text samples to faithfully explain model reasoning. This paper details the iterative design study and the resulting innovations featured in SemLa. The final design allows contrastive analysis at different levels by unearthing lexical and conceptual patterns including biases and artifacts in data. Expert feedback on our final design and case studies confirm that SemLa is a useful tool for supporting model validation and debugging as well as data annotation.


\end{abstract}  
\section{Introduction}

In natural language processing (NLP), text classification is widely used for language understanding tasks such as sentiment analysis, intent recognition, and occupation classification~\cite{suresh-ong-2021-negatives, Casanueva2020, eberle-etal-2023-contrast-bios}. For these tasks, NLP practitioners commonly adopt deep learning models like CNNs, LSTMs and pre-trained large language models (LMs)~\cite{li-2022-survey-text-classification}. Although these models score high in performance metrics like accuracy, they are well-known to be difficult to interpret and trust. As trustworthiness is crucial to practical applications, many existing visual analytics (VA) tools and explainable AI (XAI) techniques~\cite{li-2022-deepnlpvis, Lertvittayakumjorn2020, Ribeiro2016-LIME, vig-2019-multiscale-bertviz} aim to simplify the process of analyzing deep learning models.  

However, the complexity of text classification tasks has significantly increased in recent years with  more numerous and more nuanced labels becoming increasingly common~\cite{Casanueva2020, mekala-etal-2021-coarse2fine}. In such fine-grained text classification tasks, datasets have complex semantic structures comprising intricate interconnections between the many labels, and fine-grained understanding of label meaning is required to distinguish between similar labels~\cite{suresh-ong-2021-negatives}. Due to these additional challenges, existing VA tools struggle to meet the requirements of fine-grained text classification (Section~\ref{section:related_work}). 

Motivated by this gap, we introduce our novel VA system SemLa, which is designed in an iterative design-and-evaluation process in close collaboration with NLP experts from both industry and academia. Building on our previous project on interpretable fine-grained text classification~\cite{battogtokh-fine-grained-2023}, we started our collaboration on this project in late 2022 and worked together to identify and tackle the challenges in the workflow of developing fine-grained text classification models in practice. We developed and evaluated our system iteratively through multiple rounds of expert feedback each followed by improvements to the system.

Our final system supports the streamlining of various tasks in model development workflow, as we demonstrate through case studies and validate via expert feedback. The capabilities of our system include showing discrepancies between ground-truth data distribution and what the model has learned, unearthing lexical and conceptual patterns including biases from data, sample-level explanations that explicitly show fine-grained label semantics, and label-level insights that help understand relationships between different classes or within the same class. Our contributions in this paper are as follows:
(i) The design of our visual analytics system SemLa (\textbf{Sem}antic \textbf{La}ndscape) (and its components) for fine-grained text classification;
(ii) Documentation of the iterative design study, including reflections and discussion; and
(iii) Detailed evaluation of the final design based on expert feedback and case studies.

\section{Related Work}\label{section:related_work}
\subsection{Visual Analytics for Deep Learning Models in NLP}

There exists an extensive body of work leveraging VA systems and techniques to understand, assess, and debug deep-learning models in various domains including NLP~\cite{PoloChau_2019_survey_dl, la-rosa-survey-2023, gou-2021-vatld, thilo-2020-explainer}. Earlier related works include individual techniques that focus on sample-level (local) explanations such as saliency visualization of encodings for understanding phrase composition~\cite{li-etal-2016-visualizing}, bipartite-graph visualization for understanding the behavior of self-attention in Transformer-based language models~\cite{vig-2019-multiscale-bertviz}, or saliency visualizations of feature importance based on model-agnostic black-box explanation techniques like LIME~\cite{Ribeiro2016-LIME} and SHAP \cite{Lundberg2017-SHAP}.

Other works support discovery and extraction of label-level insights. For example, FeatureInsight \cite{michael_2015_viz_summary} allows users to inspect model errors of a specific label by comparing two groups of samples by the words that are most unique to each group. Similarly, FIND \cite{Lertvittayakumjorn2020} visualizes the words that a model has learned to associate with a class using word clouds and involves humans in the loop to assess flaws in the model's associations. 

Most similar to our work is DeepNLPVis \cite{li-2022-deepnlpvis}, a VA tool which supports analysis of a wide variety of deep learning models through corpus-level, word-level, and sample-level views. However, it has limited usability for fine-grained text classification. Large number of classes remain a challenge since its corpus-level view requires users to iteratively select only two labels at a time, which means understanding all key inter-relationships between many labels is infeasible due to combinatorial explosion. 
The sample-level visualization does not consider fine-grained label semantics, essential for understanding similarities and differences between analogous labels \cite{mekala-etal-2021-coarse2fine, luo-etal-2021-dont-miss-labels, battogtokh-fine-grained-2023}.

\subsection{Text Corpora Visualization}
As our work aims for multi-level analysis including corpus level, text corpora visualization (which focuses on visualizing datasets rather than models) is also relevant to us.
The closest approach in this category is topic modeling, which is the task of identifying high level topics in a corpus. 
Traditional topic modeling approaches use topic models such as Non-negative Matrix Factorization (NMF) \cite{Lee1999-nmf} and Latent Dirichlet Allocation (LDA) \cite{lda-2003}. Another category of related works focus on \textit{hierarchical} topic modeling. These include Semantic Concept Spaces \cite{assady-2019-semcon}, ArchiText \cite{architext-2021} and TopicBubbler \cite{TopicBubbler-2023}. Unfortunately, all of the above approaches are only applicable for analyzing datasets and are inapplicable for analyzing text classification models.

\section{Domain Background}\label{section:background}

In this section, we detail domain background comprising  related XAI literature and the  workflow of the collaborating NLP experts,
essential to anticipate practical challenges and requirements described in Section~\ref{subsec:requirements}.

\subsection{Explainable AI}\label{subsec:xai}

\paragraph*{Contrastive explanations}\label{subsub:contrastive_exp} Contrastive explanations answer ``Why P rather than Q?'' \cite{jacovi-etal-2021-contrastive, Miller2019}, a natural question when a model makes an error. These explanations aim to reveal the ``distinguishing factors'' that separate one outcome from the other, and are more suitable for explaining differences between similar labels (than 
simple feature importance estimates \cite{Ribeiro2016-LIME, sundarajan-2017-integrad}). Existing works \cite{jacovi-etal-2021-contrastive, ross-etal-2021-explaining} adopt a minimal approach, which aims to identify the most distinguishing factor. The minimal approach does not elaborate \textit{why} a word is considered to be distinguishing, as it fails to show how the word is related (or unrelated) to each outcome (label). In an ideal explanation, the labels should be represented by fine-grained aspects such that the explanation shows how (through which aspects) a word is related to a label \cite{battogtokh-fine-grained-2023}.

\paragraph*{Evaluating explanations}\label{subsubsec:evaluating_explanations}
As recent work incorporating multiple XAI methods into a VA framework acknowledged, ``trust in the explanation methods themselves'' remains a significant issue \cite{thilo-2020-explainer}.
``How to evaluate explanations?'' is still an open question. Related XAI studies evaluate explanations on multiple different aspects including faithfulness and plausibility, which refer respectively to how accurately an explanation represents the actual reasoning process of a model and how believable an explanation is to a human \cite{jacovi-goldberg-2020-towards}. A plausible but unfaithful explanation (which can mislead to wrong actions) should not be trusted. Therefore, faithfulness is crucial for trustworthiness. 
Several metrics have been proposed to measure faithfulness. These include measuring the impact of perturbing or erasing important words \cite{DeYoung2020,jain-wallace-2019-attention}, or guessing back the model predictions based on explanations \cite{liu-etal-2019-towards-explainable, nguyen-2018-comparing, battogtokh-fine-grained-2023}. However, there is yet no consensus on which method is the best. Importantly, as long as the question of ``How to evaluate explanations?'' remains open, any explanation method on their own cannot be fully trusted, especially if multiple competing methods are in disagreement with each other.

\subsection{Model development cycle}\label{subsec:workflow}
\begin{figure}[t]
    \centering
    \includegraphics[width=0.47\textwidth]{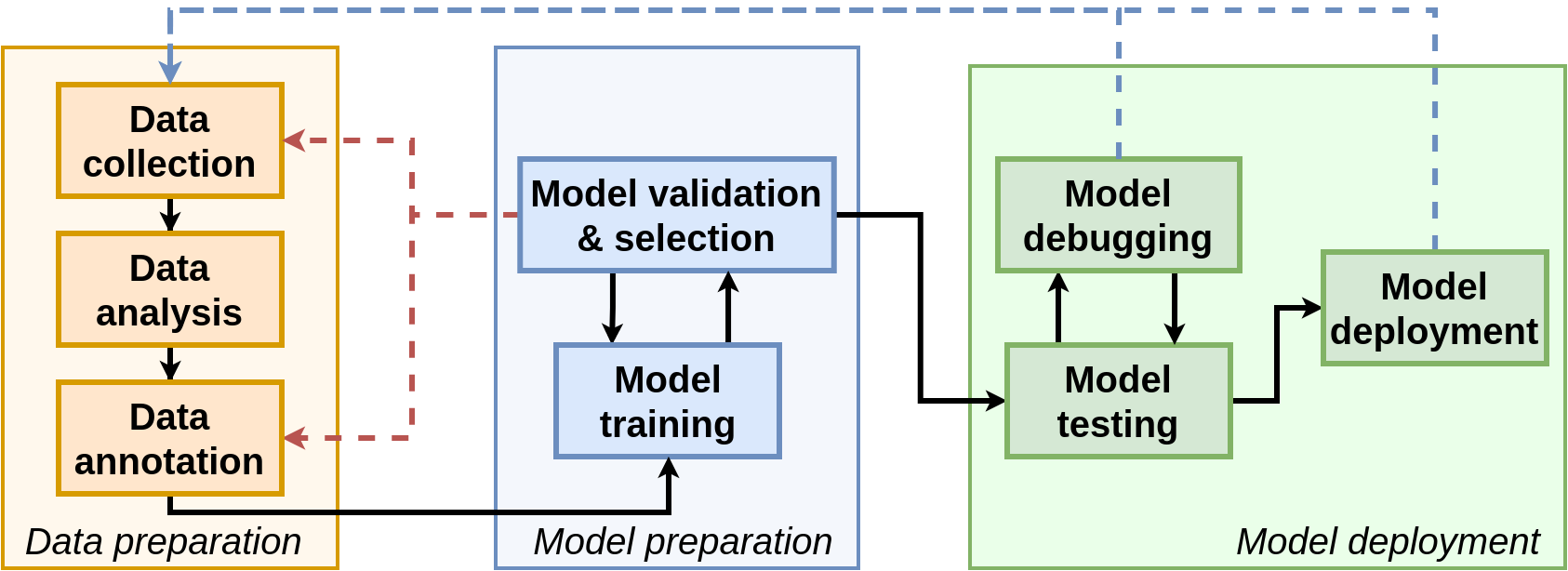}
    \caption{Model development cycle.}
    \label{fig:pipeline}
\end{figure}

To understand the practical problems experts are facing in their daily workflow, we characterized their \textit{model development cycle} based on detailed discussions and preliminary questions in our evaluation rounds (Section~\ref{sec:evaluation}). The cycle, which is illustrated in Fig.~\ref{fig:pipeline} in a simplified form that only includes the most important transitions between the stages,
consists of three main phases: data preparation, model preparation, and model deployment. 

The data preparation phase in turn consists of three stages: data collection, analysis, and annotation. User data is first collected from conversations in the \textit{data collection} stage and then manually analyzed for patterns so as to design the label set in the \textit{data analysis} stage, which involves the experts using visualization techniques like topic modeling and dimension reduction. Each sample is then assigned a label in the last \textit{data annotation} stage of this phase.

During training, it is common to save model checkpoints and train multiple models with different hyperparameter settings. In the next \textit{model validation and selection} stage, the experts evaluate the models using validation data and standard performance metrics (e.g., accuracy), and select a model. This takes substantial manual effort involving ad-hoc applications of visualization and XAI techniques (e.g., feature importance estimation, dimension reduction) using tools like Jupyter Notebook, Holoviews, 
and spreadsheets.
As shown by the red dotted arrows in Fig.~\ref{fig:pipeline}, the experts often go back at this stage to the data preparation phase and do more data collection or \textit{re-annotation} (e.g., upon identifying issues with the data annotation like simple mislabels or reasons to merge multiple labels into one, or to separate one label into multiple).

Finally, once a model is trained,  the experts debug the model before deployment, which entails similar manual efforts as during validation. 
As shown by the blue dotted arrows in Fig.~\ref{fig:pipeline}, it is normal practice to collect data after model debugging or even deployment to continually improve the model.

\section{Requirements Analysis}\label{subsec:requirements}
This section establishes the domain requirements and corresponding visual tasks, distilled from analysis of existing visualization literature (Section~\ref{section:related_work}), domain background in XAI (Section~\ref{subsec:xai}), and iterative discussions with the domain experts and understanding of their workflow (Sections~\ref{subsec:workflow} and \ref{sec:designstudyoverview}). 


\subsection{Domain requirements}

Five key requirements were identified in our investigation:

\begin{enumerate}[noitemsep,nolistsep,leftmargin=*]
    \item[R1] {Hierarchical understanding of model reasoning.} 
    Achieving a multi-level understanding of model predictions and errors, using high-level (general) and low-level (specific) insights.

    \item[R2] {Fine-grained explanation of individual predictions.} 
    Explaining the fine-grained aspects that distinguish similar labels.

    \item[R3] {Revealing model weaknesses.} 
    Uncovering weaknesses like spurious features, biases, and root cause of frequent confusions.

    \item[R4] {Visualizing semantic characteristics of labels.} Understanding the inter-relationships between labels, dissecting the fine-grained concepts, and identifying sub-clusters within a label.

    \item[R5] {Safeguarding against false impressions.} 
    Faithfully representing model reasoning with minimal built-in assumptions and supporting users in critically assessing the explanations to avoid misconceptions.
\end{enumerate}


\subsection{Tasks}
Abstracting tasks is a crucial step in translating domain requirements into actionable elements within the visualization system. By meticulously analyzing the domain requirements, we have identified a set of key tasks. To satisfy the requirement of enhancing the hierarchical understanding of model reasoning (R1), the tasks are categorized based on the level of investigation they cater to. 

%
%
%

\begin{enumerate}[noitemsep,nolistsep,leftmargin=*]
    \item \textbf{Global-level:}
    \begin{itemize}[noitemsep,nolistsep,leftmargin=*]
    \item [T1] Identifying high-level patterns in encoding space: 
    Tied to the requirement for hierarchical understanding 
    (R1), this task focuses on model validation and identifying systemic patterns.
    \item [T2] Identifying areas of weakness: Directly addressing the requirement to reveal model weaknesses (R3), this task involves pinpointing weaknesses for model validation and debugging.
    \item [T3] Comparing models: Aligned with the requirement for hierarchical understanding (R1), this task involves high-level comparisons between models.
    \end{itemize}
    
    \item \textbf{Label-level:}
    \begin{itemize}[noitemsep,nolistsep,leftmargin=*]
    \item [T4] Explaining and highlighting decision boundaries between labels: Tied to the requirement for understanding the characteristics of labels (R4), this task is vital for debugging and understanding how labels relate to each other.
    \item [T5] Explaining how a model (mis)understands a certain label: This task supports revealing model weaknesses by identifying misconceptions in label understanding (R3).
    \item [T6] Identifying similarities between label groups: Aligned with the requirement to visualize semantic characteristics of labels (R4), this task aids in debugging and data re-annotation.
    \item [T7] Identifying sub-clusters within one label group: Relating to the same requirement (R4) as the previous task, this task aids in debugging and data re-annotation.
    \end{itemize}
    
    \item \textbf{Sample-level:}
    \begin{itemize}[noitemsep,nolistsep,leftmargin=*]
    \item [T8]     
    Explaining the importance of each word through multiple metrics: This task, crucial for model debugging, aligns with explaining individual samples (R2) and safeguarding against potential false impressions from a single metric (R5).
    \item [T9] Providing fine-grained contrastive explanations: Supporting the requirement for fine-grained sample-level explanations (R2) and safeguarding against potential false impressions from the coarse-grained feature importance explanation (R5), this task focuses on analysis for debugging and validation.
    \end{itemize}
\end{enumerate}


    

\section{Iterative Design Study}\label{sec:designstudyoverview}
\begin{figure*}[t]
    \centering
    \includegraphics[width=\textwidth]{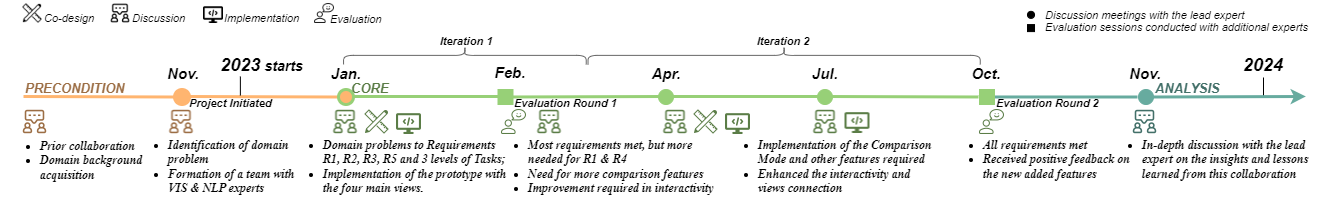}
    \caption{Timeline of the iterative design process.}
    \label{fig:timeline}
\end{figure*}

The SemLa VA system, emerged from an intensive one-year design study, engaging with six NLP experts. Their diverse expertise and roles, along with their specific contributions at various stages of the design study, are detailed in Table~\ref{tab:UserBackground}. Adhering to the structured Nine-Stage Design Study Framework~\cite{sedlmair2012design}, we ensured a methodical construction of the system,
which embraced an iterative approach encompassing cycles of design, implementation, and evaluation. In this section, we delineate the conduct of our design study, detailing the chronological progression of the iterative development process as depicted in Fig.~\ref{fig:timeline}.

\subsection{Precondition Stages} 
In Nov. 2022, our journey began with an initial consultation with an industry NLP expert (E1) from an AI research and development team. During the meeting, there was recognition of the significant potential of visual analytics in enhancing the model development workflow within the NLP domain. This consensus laid the groundwork for our dedicated exploration in this direction. Previously, we had engaged in a productive collaboration with the same leading expert, focusing on research in the domain of explicating fine-grained text classification \cite{battogtokh-fine-grained-2023}. Leveraging this domain knowledge gained from the prior engagement, our research endeavors now expanded to include visual analytics, guided by a thorough and targeted literature review. Driven by the aim to overcome the challenges posed by existing VA systems, particularly their inadequacies in managing the combinatorial complexity associated with fine-grained text classification featuring numerous labels (motivation behind R1), we conceptualized an approach. The strategy involved spatializing the samples of all labels within one model embedding space, illustrating the relationships among labels through the spatial distribution and proximity of sample points. This approach can make the relationships between similar labels and samples visually discernible, and in combination with a high-level of interactivity, enable users to navigate and explore this space–- encompassing diverse neighborhoods and areas–- at various levels of granularity.
Driven by this interactive spatialization approach, we implemented an initial proof-of-concept system. This rudimentary version comprised the early forms of two key components: the \textit{\textbf{Map}} view (Fig \ref{fig:teaser}a) prototype and the \textit{label-cluster list} (Fig \ref{fig:teaser}g) within the \textit{\textbf{Label-level}} view (see Section \ref{subsec:system-overview}). 

\begin{table}[!t]
 \centering
 \caption{Summary of the interview participants' background.}
 \label{tab:UserBackground}
 \scalebox{0.85}{
 \begin{tabular}{p{0.02\columnwidth}p{0.36\columnwidth}p{0.3\columnwidth}p{0.24\columnwidth}}
 \toprule
  \textbf{\#} & \textbf{Domain expertise}  & \textbf{Role}  & \textbf{Involvement} \\ \hline
  \textbf{E1} & 5 years in NLP & AI team leader & Entire study\\  \hline
  \textbf{E2} & 7 years in dialogue system & Client liaison \& Model development & Evaluation 1 \& 2\\  \hline
  \textbf{E3} & 7 years in NLP & Model development \& data analysis & Evaluation 1\\ \hline
  \textbf{E4} & 3 years in NLP \& dialogue system & Configuring generic models to client specifications & Evaluation 2\\ \hline
  \textbf{E5} & 2 years in NLP & PhD in abusive language detection & Evaluation 2\\ \hline
  \textbf{E6} & 2 years in medical NLP & PhD in verification & Evaluation 2\\ 
\bottomrule
 \end{tabular}%
 }
 \end{table}
 
\subsection{Core Stages: Iteration 1} 

\underline{\textbf{\textit{Discussion:}}} 
In Jan. 2023, the preliminary version of the system was showcased to the lead expert during a meeting. The consensus was that the spatialization approach showed promise in tackling R1 and could serve as a foundational element in the system. However, it was also recognized that the current design fell short of fully meeting the domain experts' needs, indicating the necessity for additional features and integrations. Following an in-depth discussion about the domain-specific challenges and the expert's vision for the tool's functionality, a set of requirements was delineated, primarily focusing on enhancing the explanation of individual predictions (R2) and the identification of model weaknesses (R3).

\noindent\underline{\textbf{\textit{Co-design \& Implementation:}}}
Prioritizing the crucial need to expose model weaknesses (R3), we introduced features such as error filtering and a confusion table to the leading expert. 
We opted for presenting the confusions in a table format (Fig. \ref{fig:teaser}h), which offered compactness and clarity as well as the ability to sort confusions based on frequency through simple interactions with the table header. 
Following this, we developed the \textit{\textbf{Sample-level}} view (Fig.~\ref{fig:teaser}c-f), directly addressing the requirements for elucidating individual predictions (R2) and pinpointing model weaknesses (R3). 

Initially, our focus was on integrating a straightforward feature importance functionality. However, given the imperative to avoid false impressions (R5) and taking into account insights from the domain background (Section~\ref{section:background}), it became evident that incorporating results from various metrics was essential. 
Consequently, we crafted our Visually Integrated Feature Importance (VIFI) view (Fig.~\ref{fig:teaser}d), which introduces minimal complexity yet provides a user-friendly interface to investigate the significance of words according to multiple diverse metrics. We soon reacknowledged that while feature importance is informative, it inherently offers a coarse-grained perspective. To address this, we turned our attention to contrastive explanations and observed that existing works often leave label semantics implicit, whereas making these explicit is more advantageous \cite{battogtokh-fine-grained-2023}. In response, we conceptualized an example-based contrastive explanation approach, incorporating three distinct visualizations (Fig. \ref{fig:teaser}c,e,f), detailed in Section~\ref{subsub:relcharts}.

\noindent\underline{\textbf{\textit{Evaluation:}}}
The current prototype, designed to meet requirements R2, R3, and R5, was evaluated by three industrial experts E1, E2, and E3 through semi-structured interviews, detailed in Section~\ref{sec:evaluation1}. While receiving positive feedback, experts also highlighted areas for enhancement, including the need for 1) a model comparison feature, 2) clearer headers in sample view relation charts, and 3) improved bidirectional and hierarchical sorting of the confusion list. 




\subsection{Core Stages: Iteration 2}

\underline{\textbf{\textit{Discussion:}}} 
After reviewing the interview feedback with the lead expert E1, we identified two additional key requirements based on the suggestions received: 1) the need for a hierarchical understanding that encompasses high-level model comparisons (R1), and 2) the necessity for more in-depth comparative analysis of label similarities and differences (R4) both between and within groups.

\noindent\underline{\textbf{\textit{Co-design \& Implementation:}}}
In response to the feedback from Iteration 1, several improvements were integrated, e.g., headers were added to the relation charts for better clarity and the confusion table received hierarchical sorting capabilities. To meet the requirement of R1 \& R4, a new abstraction layer was introduced for local words, enabling the visualization of localized commonsense concepts. Furthermore, a \textit{comparison mode} feature (Fig~\ref{fig:teaser}i-j) was implemented, providing the capability to contrast any two groups of samples. This mode was compatible with the system's existing functionalities, offering options such as concept-based filtering, ground-truth vs. prediction comparisons, label differentiation, a lasso tool, and sample-level analysis for longer texts.

\noindent\underline{\textbf{\textit{Evaluation:}}}
In Oct. 2023, upon the completion of the system, a second round of evaluation was initiated. This phase involved five domain experts: two returning from the initial round (E1 \& E2), a new member from the same industrial AI team (E4), and two additional evaluators (E5 \& E6) with an academic background encompassing NLP, XAI, and VIS. Employing a similar approach as before, semi-structured interviews were conducted with each evaluator. The detailed procedure of this evaluation, along with the insights and feedback gathered, is thoroughly documented in Section~\ref{sec:evaluation2}. The follow-up survey revealed a unanimous endorsement of the system by the evaluators. Feedback was uniformly positive across all components, validating the fulfillment of all predefined requirements. Particularly, the \textit{comparison mode} garnered strong recognition for its practicality and effectiveness.

\subsection{Analysis Stages}
Post the second evaluation round, we gathered comprehensive feedback from the domain experts, leading to in-depth discussions and reflections on the lessons learned during the design study. These valuable insights and analyses are discussed in detail in Section~\ref{sec:discussion}. 










\section{SemLa: System Description}

In this section, we describe our novel VA system \textbf{SemLa} and our innovative visualization techniques.

\subsection{Overview}\label{subsec:system-overview}
The system interface consists of four coordinated views. The \textbf{Map} view projects a corpus to a 2D scatter plot using sample embeddings by a selected model and dimension reduction (Fig.~\ref{fig:teaser}a). 
Each sample is represented by a circle (or another shape at low-level to differentiate labels). At the high level, as there is an excessive number of labels to visually encode, labels are grouped into clusters, which are encoded in the colors of the samples. 
The Map view is equipped with various features for dissecting the visual space, including zooming, panning, filtering, and a novel interactive \textit{local word} visualization (see Section \ref{subsub:local_clouds}), which helps users navigate and understand the semantic contents of different neighborhoods. The \textbf{Lists} view (Fig.~\ref{fig:teaser}b) synchronously provides a ranked summary of the concepts, words, and labels present in the samples currently visible on the Map view as the user interacts with the system. Both of these views also have a \textit{comparison mode} (Fig. \ref{fig:teaser}i-j) option for contrastive analysis, in which the Map view presents two separate scatter plots (that can show two different groups of samples, or the same corpus through two different models) and the List view shows which concepts, words, and labels are equally likely in both groups and which ones are more likely to be in one group than in the other. The \textbf{Sample-level} view explains a single prediction upon the user selecting an individual sample (Fig.~\ref{fig:teaser}c-f). This is done through four visualizations (see Sections~\ref{subsec:vifi} and \ref{subsub:relcharts}). 
Lastly, the \textbf{Label-level} view consists of \textit{label-cluster list} (Fig.~\ref{fig:teaser}g), which lists labels grouped by similarity, and \textit{confusion table} (Fig.~\ref{fig:teaser}h), which is a table showing which pairs of labels are most frequently confused. It is possible to sort this table by its columns including confusion frequency, which means users can effortlessly identify which labels were most frequently mistaken for one another. 
Furthermore, users can select one or more labels from the label-cluster list or confusion table to investigate those labels on the Map and List view.

\subsection{Local Words}\label{subsub:local_clouds}

For complex semantic structures consisting of many neighborhoods with hierarchical relations, a method for identifying patterns at multiple levels is necessary. 
The closest existing work for this task is BERT-based topic modeling \cite{topic-models-vis-2023}, which is applicable not only for explaining datasets but also models. It depends on clustering to identify topics in a top-down manner by first grouping samples into clusters and then extracting the top keywords (using class-based \textit{tf-idf} analysis \cite{grootendorst2022bertopic}). The problem with this dependency is two-fold: 1) clustering is computationally expensive and 2) the top-down approach forces the data to be seen through an extra lens (by introducing arbitrary assumptions about data in the form of hyperparameters, e.g., the number of clusters or their density).

Therefore, we propose an alternative approach based on our simple and fast Localized Word Clouds (LWC) algorithm (Algorithm~\ref{alg:local_words}), which extracts patterns directly from the model embedding space in a bottom-up manner without relying on clustering. The intuition of LWC is to identify words that are \textit{localized} to a certain neighborhood, i.e., appearing frequently there but not anywhere else. We do this by computing the locality of each word (the area enclosing all occurrences of the word) and allowing users to filter the words by their frequency and the sizes of their localities.

The result of using LWC is a visualization resembling a word cloud. However, it differs from traditional word clouds in that it can be overlaid meaningfully on the 2D scatter plot of the corpus. 
Over an approach employing traditional word clouds, which would require grouping the corpus into partitions and generating multiple word clouds (as in FIND~\cite{Lertvittayakumjorn2020}), our approach has the advantages of being space efficient and avoiding word repetitions. Furthermore, after representing the output words with sets of concepts, LWC can be applied recursively to extract more abstract localized concepts from the inital local words.

\RestyleAlgo{ruled}
\setlength{\textfloatsep}{15pt}
\begin{algorithm}[t]
\setstretch{0.5}
\smaller
\caption{\vspace{5px} Localized Word Clouds (LWC)\\
\smaller
LWC outputs a set of $l$ \textit{local words} $L = \{w_1, w_2, \ldots, w_l\}$ and their corresponding positions $P_L = \{ p_1, p_2, \ldots, p_l \}$ in a space $S$ (of arbitrarily many dimensions) given inputs that include $M$ samples $D = \{x_1, x_2, ..., x_M\}$ and their positions $P_D = \{p_1, p_2, \ldots, p_M\}$ in $S$. The output words are filtered by their frequency with parameter $T$ and by locality size with function $R(\cdot)$. A function  $C(\cdot)$ computes the center of the locality of word $w$, on which $w$ is to be visualized. Our choices of functions $R$ and $C$ are described in supplementary materials.
}\label{alg:local_words}
\SetKwInOut{Input}{input}
\SetKwInOut{Output}{output}

\Input{ $D = \{x_1, x_2, ..., x_M\}$, $P_D= \{ p_1, p_2, ..., p_M \}$, R, T, C}
\Output{ $L = \{w_1, w_2, ..., w_l\}$, $P_L= \{ p_1, p_2, ..., p_l \}$ }

$L \gets \{\}$\;
$P_L \gets \{\}$\;

$W \gets $ an empty map\;

\For { $x_m \in D$ } {
    \For { $w_j \in x_m$ } {
        \If { $w_j \notin W.keys()$} {
            $W[w_j] \gets $ an empty list\;
        }
        $p \gets p_m \in P_D$\;
        $W[w_j].add(p)$\;
    }
}

\For {$w_i \in W.keys()$} {
    $( p_1, p_2, ..., p_F ) \gets W[w_i]$\;

    \If { $R( p_1, p_2, ..., p_F)$ and $F > T$} {
        $p_i \gets C(p_1, p_2, ..., p_F) $\;
        $L.add(w_i)$\;
        $P_L.add(p_i)$\;
    }    
}

\Return $L, P_L$
\end{algorithm}

\subsection{Visually Integrated Feature Importance (VIFI)}\label{subsec:vifi}

Our Visually Integrated Feature Importance (VIFI) view seamlessly integrates multiple feature importance metrics into one simple visualization (Fig.~\ref{fig:teaser}d). Each bar segment in the chart corresponds to a different feature importance metric. This design makes it simple to see 1) how much importance in total a word receives from different metrics and 2) how much each metric contributes.

\subsection{Example-based contrastive explanations}\label{subsub:relcharts}

To achieve fine-grained explanations that make the label semantics explicit (see Section~\ref{section:background}), we adopt a novel approach that combines example-based explanations with contrastive explanations. By representing a label with in-distribution samples of that label, which exemplify its fine-grained aspects, our explanations explicate why certain words may or may not relate to the label. Combining this concreteness with contrastiveness, our explanations detail why and how the words in a given sample relate to two different labels. We incorporate three visualizations of this approach in SemLa, all of which explain a query sample with respect to two other samples: the closest sample, which shares the same label as the query sample, and a contrast sample, which has a different label. Out of our three visualizations, one shows natural language summary (Fig.~\ref{fig:teaser}c), and the other two (which we refer to as \textit{relation graphs} collectively) show token-to-token links between the three samples (Fig.~\ref{fig:teaser}e) and the contribution of each token to the similarities computed between the three samples (Fig.~\ref{fig:teaser}f).

\section{Evaluation}\label{sec:evaluation}

Throughout our iterative design study, two structured evaluation rounds were conducted: an initial session in Feb 2022, assessing the tool's alignment with the primary requirements, followed by a subsequent session in Oct 2023, which focused on evaluating the overall usability of the final tool and specifically examining the enhancements made in response to feedback previously received.
Through the evaluation-related activities, several representative cases emerged from in-depth discussions with the experts and observation of their behavior. These cases are particularly focused on model debugging and are presented in Section~\ref{section:case-studies}.
%

\subsection{Evaluation Round 1:  Feb. 2022}\label{sec:evaluation1}
\subsubsection{Protocol}
\textbf{Time and Participants.}
This round of evaluation conducted in Feb. 2022, involved the participation of three dialogue system experts from the industry, specializing in delivering conversational AI solutions to client companies, which subsequently utilize these solutions to engage with end customers. Participants E1 and E3 are deeply involved in model development, while E2 assumes a more client-centric role. A unifying aspect of their professional roles is a keen interest in elucidating model reasoning and pinpointing model weaknesses, integral tasks to their daily responsibilities.

\noindent\textbf{Activities and Duration.}
The evaluation encompassed three individual semi-structured interviews, each tailored to assess the system at its initial stage. Key components under review were the \textbf{\textit{Map}}, \textbf{\textit{List}}, \textbf{\textit{Sample-level}}, and \textbf{\textit{Label-level}} views (Fig. \ref{fig:teaser}). Following a standardized format, outlined in Table~\ref{tab:procedureandduration}, each interview spanned approximately ninety minutes.
Sessions commenced with a preliminary preparation during which objectives were clarified. 
A succinct overview of the tool, highlighting its main features and a demonstration using the BANKING77 public dataset \cite{Casanueva2020}, known for its fine-grained intents, was provided. This choice was also influenced by privacy considerations associated with using actual user data. The system incorporates a BERT model \cite{devlin-etal-2019-bert}, trained via metric-based representation learning method \cite{chen-2022-contrastnet} for 5-way 1-shot classification. This configuration facilitates explaining predictions via distances (from the query sample to a support set comprising samples from five distinct labels, with one sample per label).
Subsequently, experts engaged with SemLa, experimenting with its functionalities across four predefined tasks (see supplementary material). The tasks, designed to mirror specific requirements and objectives, provided insights while experts verbalized their thoughts. Our focus was on assessing the system's capabilities and the efficacy of each feature in meeting the established requirements.
The evaluation concluded with a reflective survey, incorporating both open-ended and closed-ended questions. The survey aimed at gathering holistic feedback, encompassing the system features' effectiveness in fulfilling domain needs, its overall usability, 
and suggestions for enhancements.

\subsubsection{Results}
The usability of SemLa, assessed via Likert-scale questions, received highly positive feedback. Experts unanimously \textit{strongly} agreed on the system's overall usefulness and its ability to clarify individual predictions and identify label sub-clusters (R2). For aspects like identifying model weaknesses (R3), high-level understanding of models, explaining decision boundaries within labels, and identifying semantic overlaps between labels, the majority \textit{strongly} agreed on the system's effectiveness, with one expert \textit{somewhat} agreeing. As for the visualizations' intuitiveness, one expert \textit{strongly} agreed, while the others \textit{somewhat} agreed. A detailed summary of the experts' responses to open-ended questions aligning with these ratings will be presented in the following paragraphs.

\textbf{Use Cases and System Utility:}
The experts expressed significant enthusiasm about the system's potential, describing it as \say{\textit{immensely valuable}} for tasks like model debugging, which involves pinpointing weaknesses and understanding the root causes of errors. Additionally, they highlighted its utility in client communications. One expert noted the system's capability to \say{\textit{intuitively spot weaknesses}} by enabling users to 1) filter errors using confidence thresholds (referred to as \say{\textit{top and tail}} the errors), and 2) comprehend these errors for \say{\textit{targeted intervention}} rather than relying on \say{\textit{trial-and-error}} approaches for model improvement.

\textbf{Most Effective Visualizations:}
The experts varied in their preferences for the system's visualizations. One expert highlighted the \textit{sample-level} visualizations, particularly the contrastive explanations, as \say{\textit{novel and very useful}.} Another expert emphasized the utility of integrating the confusion table with token-to-similarity relations, finding it instrumental in grasping the model's primary errors. The third expert valued the combination of Local Words visualization with the interactive features of the \textit{map}, appreciating its versatility in offering deeper insights.

\textbf{System Novelty and Capabilities:}
Responses to the system's novelty largely focused on its \textit{sample-level} explanations. All experts agreed that the system provides \say{\textit{deeper insights}} compared to existing tools at their disposal. 
The two experts (E1 \& E3) deeply involved in model development provided specific insights: one remarked they had 
\say{\textit{never seen anything like this}} that digs deep into the root causes of errors despite being experienced with explanation and visualization techniques (e.g., LIME, topic modeling); the other noted that, unlike current tools, our system enables a clearer understanding of error causation.
Particularly, they highlighted the relation graphs (Fig. \ref{fig:teaser}e-f) that offer contrastive explanations as the most innovative and useful visualizations, marking a significant advancement in model interpretability.

\textbf{Visualization Understandability and Learnability:}
Overall, all experts agreed that the visualizations were intuitive and easy to grasp with minimal learning required. They did, however, suggest some enhancements for clarity, such as adding \textit{column headers} to the relation graphs for easier sample identification,
and tooltips providing further meta-details about the visualizations (e.g., explanation of value calculations). 
One expert particularly noted the system's compatibility with their existing workflow, stating it would integrate seamlessly and enhance their work process. They emphasized the system's ability to easily reveal insights that are \say{\textit{very difficult to find from spreadsheets}} (raw tabular data), thereby significantly improving their workflow experience.


\textbf{Recommendations for System Enhancement:}
Experts suggested that our system could be more effective in \textit{\textbf{facilitating comparisons between different models}}, such as checkpoints or models resulting from different runs with varied hyperparameters. While acknowledging the system’s current capacity to support model comparison by loading different models, they recommended features specifically tailored for this task. Their suggestions include a sequential time-lapse animation depicting changes in the semantic landscape across model checkpoints, or a parallel view comparing two distinct models. 
Other recommendations reiterated earlier suggestions, such as adding tooltips to provide meta-details about the visualizations and implementing column headers in the relation graphs for enhanced clarity.

 \begin{table}[!t]
     \centering
     \caption{Interview Procedure and Duration.}
     \label{tab:procedureandduration}
     \scalebox{0.9}{
     \begin{tabular}{ccc}
     \hline
      \textbf{Order of Procedure}                    &\textbf{Activities}                                    & \textbf{Duration}  \\ 
     \hline
      {\makecell[c]{Preliminary\\ Preparation}}      & {\makecell[l]{1) Introductory questioning \\ 2) Tool walkthrough}}     & 10-15 min  \\ 
     \hline
     {\makecell[c]{Task Scenarios}}                 
     &  1) Test via predefined tasks                                             & 45-60 min \\ 
     \hline
     {\makecell[c]{Follow-up\\ Survey}}       
     & {\makecell[l]{
        1) Reflection on the tool \\ 
        2) Likert-scale questions
     }}       & 15-20 min \\ 
     \hline
     \end{tabular}%
     }
     \vspace{-0.05in}
 \end{table}

\subsection{Evaluation Round 2: Nov. 2023}\label{sec:evaluation2}
Building on the insights gathered from the previous evaluation, SemLa's features underwent further refinement, adhering to the iterative process. The usefulness and usability of the tool were then re-evaluated through one-on-one interviews, employing the think-aloud protocol.
During these sessions, we closely observed the participants' interactions and comments about the tool, conducting a qualitative analysis based on the interview transcripts and data from the follow-up survey.

\subsubsection{Protocol}
\textbf{Time and Participants.}
In Nov. 2023, the second evaluation round was conducted, broadening its reach to include domain experts from both industrial and academic sectors. This approach aimed to gather extensive and unbiased feedback about the system. Five participants were involved: E1 and E2, who had participated in the previous evaluation, were joined by E4, a colleague from their industrial AI team. Academia was represented by E5 and E6, experts with two years of NLP experience each in both academic and industrial settings. The evaluators' familiarity with NLP, XAI, and VIS was gauged through background questions in the survey, acknowledging that varied domain expertise could shape their perceptions of the system.
Each of the five participants had NLP experience, averaging 3.8 years in the field. All five rated their familiarity with XAI as moderate to high, and similarly, they were well-versed in visualization techniques. This assortment of expertise provided a more comprehensive backdrop for assessing the system’s efficacy and usability across diverse domain backgrounds.

\textbf{Activities and Duration.}
Mirroring the methodology of the previous evaluation, we continued with semi-structured interviews, adhering to the specific process and timing detailed in Table~\ref{tab:procedureandduration}. 
To illustrate the generalizability of the system, we demonstrated it on additional datasets for fine-grained text classification, which include Medical Bios \cite{eberle-etal-2023-contrast-bios}, GoEmotions \cite{demszky-etal-2020-goemotions}, and HWU64 \cite{Liu2021-hwu}.
This round, while still assessing the overall effectiveness and usability of the system, placed greater emphasis on evaluating the impact of the newly integrated modes and functionalities. 
This focus was reflected in the design of the evaluation tasks, the nature of the reflective discussions, and the structuring of the survey questions.

\noindent\subsubsection{Results}

\begin{figure*}[t]
    \centering
    \includegraphics[width=0.9\textwidth]{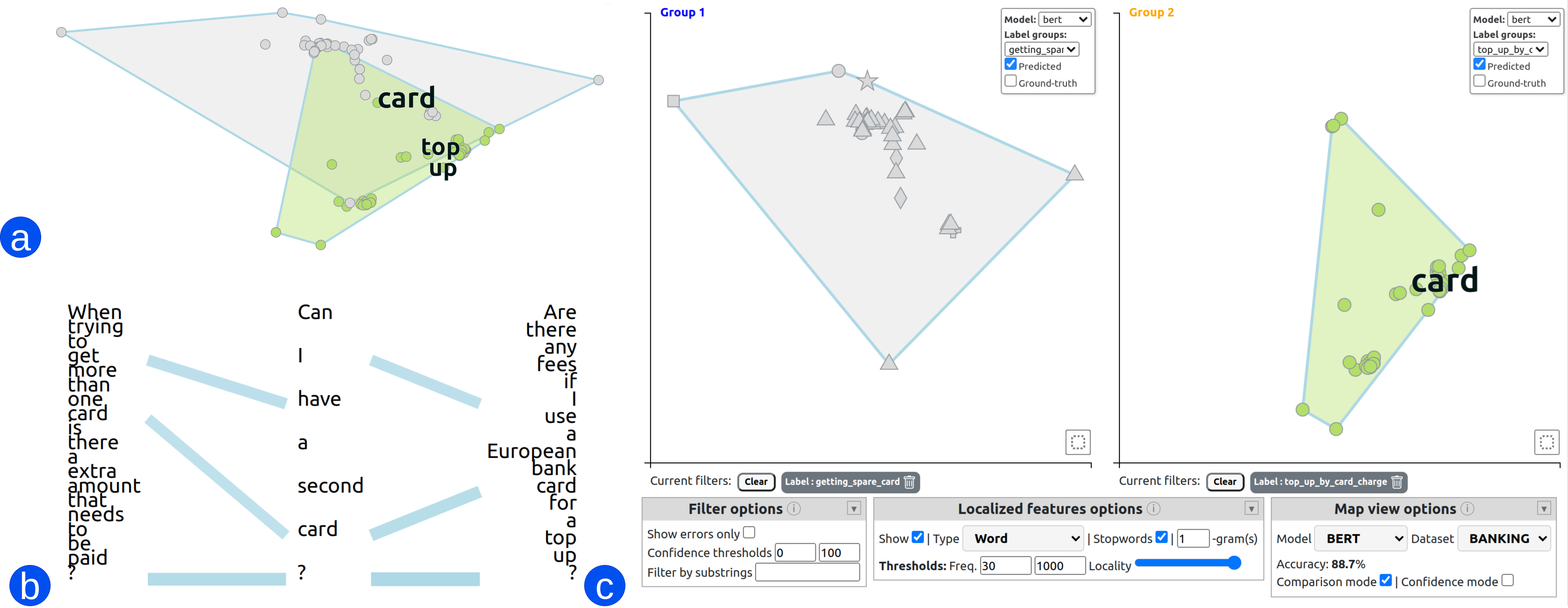}
    \caption{Multi-level analysis results: a) local words in the two top-confused labels \textit{getting\_spare\_card} and \textit{top\_up\_by\_card\_charge}, b) label \textit{topping\_up\_by\_card\_charge} has the word ``card'' more often than the label \textit{getting\_spare\_card} in the model predictions, as the word ``card'' appears over the former but not the latter at frequency threshold of 30, and c) token-to-token links in a false positive case of \textit{getting\_spare\_card} (represented by an example on the left) being mistaken for \textit{topping\_up\_by\_card\_charge} (represented by an example on the right) confirms that the word ``card'' was a confounding feature.}
    \label{fig:multi_level_analysis}
\end{figure*}


%
\textbf{Reflection on Overall Usability:}
The survey results showed the system had a good level of overall usability. 
All six participants unanimously agreed on the system's utility, with a consensus indicating that the tool was indeed beneficial. Regarding learnability, there was a strong alignment among participants, ranging from agreement to strong agreement, on the visualizations being easy to understand.  
Delving deeper, the survey employed a detailed set of questions rated on a five-point Likert scale (ranging from 0 for 'strongly disagree' to 5 for 'strongly agree') to measure the system's performance against domain-specific requirements. The compiled average scores are as follows:

The system excelled in offering a high-level model understanding (R1), facilitating the understanding of individual predictions (R2), identifying model weaknesses (R3), and aiding users to discern sub-clusters within labels (R4), all achieving the top average rating of 5;
Its proficiency in both clarifying the details of individual labels (R2) and distinguishing between similar labels (R4) was evident with an average score of 4.8.
These scores collectively demonstrate the system's adeptness in providing comprehensive insights across various levels of model behavior and predictions, reinforcing its strong usability in the domain's complex analytical landscape.

\textbf{Reflection on Usefulness:}
Thematic categorization and keyword analysis of the interview transcripts revealed that participants predominantly underscored the system's usefulness in facilitating error analysis, model validation, bias detection, and exploratory data analysis at hierarchy levels.  Notably, E6 highlighted the system's capability to unearth adversarial examples, thereby contributing to an enhanced understanding of model robustness. E1, E2, and E4 agreed on the system's substantial impact on streamlining the manual effort involved in bias detection and the articulate presentation and explanation of those findings to stakeholders. 

Regarding the efficacy of specific visualizations, the \textit{comparison mode} emerged as a standout feature, receiving accolades from five participants for its effectiveness. Simultaneously, the interactive Map view
was commended by four participants, making these two features the most lauded among the participants for their contribution to the system's analytical capabilities.

Reflecting on the influence of the system on their workflows, the consensus among participants was that SemLa's unique integration of diverse methods and interactive techniques distinguishes it from prior tools. The facility to engage in multi-level analysis was deemed crucial, particularly for augmenting model validation, 
data analysis, and expediting troubleshooting. This comprehensive integration is credited with substantially enhancing both the efficiency and efficacy of their analytical endeavors.

Participants provided valuable insights into unanticipated use cases for the tool, revealing potential areas for feature expansion. Notably, they identified its capacity to discern adversarial samples and its utility in demystifying complex AI systems for non-expert audiences, such as elucidating why solutions like ChatGPT do not supplant task-oriented dialogue systems. These unexpected applications suggest promising directions for further refinement and enhancement of the tool's capabilities.

\textbf{Expectations on Improvements}
The experts commented that SemLa is at a stage where future enhancements revolve around operationalizing the system for production.
We discuss these expected improvements in our reflections in Section \ref{sec:discussion}.


\section{Case Studies}\label{section:case-studies}

Throughout our design study, we conducted in-depth case studies on various fine-grained text classification datasets. We illustrate through two case studies how SemLa can be used to tackle important requirements and streamline the model development workflow.

\subsection{Identifying Root Cause of Model Errors on BANKING77}

BANKING77 is a popular public benchmark dataset for intent recognition, which is a task known to entail especially fine-grained labels~\cite{Casanueva2020}. This dataset contains 77 labels related to the same banking domain, which makes it very fine-grained even among intent recognition datasets and representative of how complex datasets are in practical applications. We analyzed a BERT-base model (fine-tuned on the training split of this dataset) on the test split containing 3080 samples.

To begin our analysis, we filtered errors on the Map view and inspected the label distributions in the List view. From the \textit{gold label list} and \textit{predicted label list} respectively, we see the labels ranked by their shares of the false negative and false positives predictions. For example, 3.7\% and 3.4\% of the false negatives corresponded respectively to the labels \textit{compromised\_card} and \textit{supported\_cards\_and\_currencies}, whereas 3.2\% and 3.2\% of the false positives corresponded respectively to \textit{top\_up\_by\_card\_charge} and \textit{reverted\_card\_payment?}. The \textit{confusion table} in the Label-level component,  confirmed this and  sorting the table by the columns ``ground-truth'' and ``prediction'' provided a detailed breakdown of the specific confusions. Interestingly, when sorted by confusion frequency (clicking on the frequency column's header), we found the most frequent confusion was mistaking \textit{getting\_spare\_card} for \textit{top\_up\_by\_card\_charge}, which happened three times. Our goal became understanding this confusion and why the model most frequently predicted \textit{top\_up\_by\_card\_charge} false positively. 

To understand the top confusion, we looked at the local words with frequency threshold of 20 (ignoring stop words) and found that ``card'' was in the intersected area (shown by enclosing hulls corresponding to each label) of the two labels, and that ``top'' and ``up'' were only in the area of top\_up\_by\_card\_charge. 
The most common word across the two labels was ``card'', which appeared in 66.3\% of samples (Fig. \ref{fig:multi_level_analysis}a), which is expected based on the label names. 
When we lower the freq. threshold to 5, we saw more words in the intersected area that gave new insights to the confusion, which are ``add'' and ``extra'' that could relate to ``spare'' and ``charge'' in the label names. To confirm this, we switched to visualizing local concepts rather than words, and found that the concept of ``extra'' is indeed shared between the two labels (we also found out the words ``spare'' and ``additional'' were in the intersected area by clicking on the concept). This gave us a general insight into why the two labels were close to each other in model embedding space. 

Furthermore, we looked at the three errors using the Sample level component. In all three errors, the word ``card'' was found to be a confounding factor (Fig.~\ref{fig:multi_level_analysis}b shows one of these errors), that related to both the incorrect \textit{top\_up\_by\_card\_charge label} and the correct \textit{getting\_spare\_card label}, which suggested that the model associated the word ``card'' to the former more strongly. 

To confirm this, we activated compare mode and looked at the two labels side by side and found that ``card'' indeed occurred in more samples predicted to be \textit{top\_up\_by\_card\_charge} despite there being equal number of samples for each label in the dataset (Fig.~\ref{fig:multi_level_analysis}c). This was also true against the label that was second most frequently mistaken for \textit{top\_up\_by\_card\_charge}  (\textit{supported\_cards\_and\_currencies}). 

Furthermore, when we compared the samples that were \textit{predicted} to be \textit{top\_up\_by\_card\_charge} with those that actually belonged to this label, the word ``card'' was more frequent in the former group than in the latter. On the other hand, the words ``top'' and ``up'' were less frequent in the former than in the latter. These suggested the model was giving the word ``card'' more importance than it should when predicting \textit{top\_up\_by\_card\_charge}. 

In summary, our analysis showed that the model associated the word ``card'', which was common among many other labels, too strongly with \textit{top\_up\_by\_card\_charge}, which explains why the model most frequently predicted this label false positively.

\subsection{Hidden Conceptual Relations in Multi-Domain Datasets}

\begin{figure}[t]
    \centering
    \includegraphics[width=0.35\textwidth]{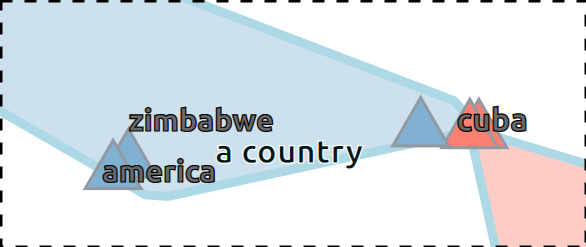}
    \caption{The concept of ``a country'' connecting the label vaccine to cancel\_reservation}
    \label{fig:vaccine-cancel_reservation-common-concept}
\end{figure}

HWU64~\cite{Liu2021-hwu} and CLINC150~\cite{larson-etal-2019-clinc} are \textit{multi-domain} intent recognition datasets with 64 labels in 18 domains and 150 labels in 10 domains respectively. We analyzed BERT-base models (each fine-tuned on the training split of the respective dataset) on the test splits containing 1076 and 4500 samples respectively. On these datasets, we often found unexpected cross-domain conceptual relationships between seemingly unrelated labels.

The most interesting case was on CLINC when looking at the most frequent confusions. The most frequent confusion was 
between the seemingly unrelated labels \textit{vaccines} and \textit{cancel\_reservation}. To investigate why, we clicked on the confusion to see the two labels on the Map view. Upon not finding apparent connection between the labels when looking at the local words, we switched to visualizing the local concepts. Then, we found there are many countries mentioned in \textit{cancel\_reservation} label (``spain", ``mexico", ``america", ``china" and ``zimbabwe"), and that the error cases all contained the word ``cuba", which is also a country. Even though ``cuba" was not among any sample that actually has the label \textit{cancel\_reservation}, the model likely made these errors after recognizing that ``cuba" was similar to other country names (Fig.~\ref{fig:vaccine-cancel_reservation-common-concept}). The system automatically extracted and instantly showed us this hidden conceptual relation, which otherwise would not have been apparent without manually looking through the data in detail. 

\section{Reflection and Discussion}\label{sec:discussion}
We identified several takeaways from our design study after analyzing our evaluation results and reflecting on our collaboration with the domain experts. 
These apply to developing a generalizable system that addresses the needs of different user profiles.

\paragraph*{Cast a wide net to generalize}
The diverse backgrounds and roles of the experts who participated in our design study entailed a wide range of requirements and individual differences in how they prioritized the requirements, as they apply text classification to different domains (dialogue system, medicine, abuse detection) and focus on different aspects (performance, explainability, robustness) . This was reflected in the expert feedback, which showed that the feature that each expert found most novel was often related to their individual background and concerns. Often, a system feature that one expert did not pay attention to was found to be the most novel by another expert. For example, the idea of our VIFI view, which was merely acknowledged by most experts, was highlighted as one of the most novel feature with strong significance by expert E5 who has high level of experience and in-depth understanding of XAI methods. Therefore, our reflection is that what may seem like unnecessary complexity to one expert can be a necessity for another. When addressing a task with wide applicability like text classification, to prevent overly tailoring our system design to only a subset of potential users, it was worthwhile to ensure that our requirements analysis encompassed not only the practical challenges experienced by the experts, but also the common problems addressed by previous works and the background domain knowledge in XAI.

\paragraph*{Resist the temptation to simplify}
Simplicity is a key factor in usability and an important design principle behind our novel visualizations. However, as previously discussed, neglecting individual differences in user requirements for simplicity would lead to poor generalizability. Furthermore, based on the questions we received from some of the experts, omitting low-level details behind the visualizations or key information that needs to be clear (e.g., how are the link strengths calculated in our relation graphs, or what are the labels that correspond to each column) can hurt transparency and reduce the simplicity \textit{experienced} by the user.

\paragraph*{Use XAI techniques responsibly}
As each individual explanation method simplifies model reasoning, each can only offer a limited perspective. Furthermore, as multiple competing methods disagree with each other (Section \ref{section:background}), acknowledging these limitations to the users and offering them multiple perspectives is essential for preventing misconceptions and using these methods responsibly. This applies to users at both ends of the spectrum when it comes to how much they know about the explanation methods and how likely they are to trust them. 
Based on our discussions with the experts, for users who lack knowledge of explanation methods, 
providing multiple perspectives and acknowledging the limitations are critical in reducing vulnerability to misconceptions, whereas for users who are generally familiar with explanation methods, the same multiple perspectives are required to address the current issue of ``trusting the explanation methods themselves''~\cite{thilo-2020-explainer}. 
This pervasive need for multiple perspectives motivates the use of visual analytics systems in practically applying XAI methods and new ways of visually integrating multiple explanations together.

\paragraph*{Accompany freedom with guidance}
In our last evaluation round, discussions with two experts pointed to a common direction for using SemLa in production, which was guiding users to intuitively follow a series of steps to complete common tasks. The experts discussed that these could be in the form of providing documentations and tutorials within and outside the system, or tailoring the default settings and the UI design of the system for the important and common usage scenarios. They further explained that the reason behind these comments is actually based on the strength of the system, which is that it offers a wide variety of features and full freedom to explore the model and the data. This resonated with the initial challenges we faced in Iteration 1, which were about how to best exploit the freedom-to-explore based on user requirements. Our reflection based on these is that while offering the users a high-degree of freedom is useful for generalizability, ensuring usability by providing guidance tailored to user requirements is essential.

\section{Conclusion and Future Work}
In this paper, we detailed the intensive iterative design study involving six NLP experts in total that resulted in our visual analytics system SemLa for analyzing fine-grained text classification models and datasets. Our evaluation of the final design based on expert feedback and case studies show that SemLa effectively addresses the special challenges posed by the task and that it can overall be a useful tool for assisting experts (with different roles and different backgrounds) in their workflow of analyzing models and datasets with a diverse range of use cases.

In future iterations, we would like to refine SemLa to be used in production scenarios by adding more ways of extracting insight from data using our LWC algorithm and assisting in communication between experts and non-experts.
We are also excited by the prospect to extend our approach based on free exploration of model embedding space to other application domains of deep learning, such as image processing and multi-modal input processing.


\bibliographystyle{eg-alpha-doi} 
\bibliography{main}       


\end{document}